\UseRawInputEncoding
\documentclass[10pt,twocolumn,letterpaper]{article}

\usepackage{cvpr}
\usepackage{times}
\usepackage{epsfig}
\usepackage{graphicx}
\usepackage{amsmath}
\usepackage{amssymb}
\usepackage{slashbox,multirow}
\usepackage{setspace}
\usepackage[pagebackref=true,breaklinks=true,letterpaper=true,colorlinks,bookmarks=false]{hyperref}

\newcommand*{\affaddr}[1]{#1} % No op here. Customize it for different styles.
\newcommand*{\affmark}[1][*]{\textsuperscript{#1}}
\newcommand*{\email}[1]{\texttt{#1}}

\cvprfinalcopy % *** Uncomment this line for the final submission

\begin{document}

%%%%%%%%% TITLE
\title{Decoupled Spatial-Temporal Transformer for Video Inpainting}

\author{%
Rui Liu\affmark[\dag]\thanks{The first two authors contribute equally to this work.} \quad Hanming Deng\affmark[\ddag]\footnotemark[1] \quad Yangyi Huang\affmark[\ddag\S] \quad Xiaoyu Shi\affmark[\dag] \quad Lewei Lu\affmark[\ddag] \\
Wenxiu Sun\affmark[\ddag$\sharp$] \quad Xiaogang Wang\affmark[\dag] \quad Jifeng Dai\affmark[\ddag] \quad Hongsheng Li\affmark[\dag\#] \\
\affaddr{\affmark[\dag]CUHK-SenseTime Joint Laboratory, The Chinese University of Hong Kong \quad \affmark[\ddag]SenseTime Research \\
\affmark[\S]Zhejiang University \quad \affmark[$\sharp$]Tetras.AI \quad \affmark[\#]School of CST, Xidian University}\\
\email{\{ruiliu@link, xiaoyushi@link, xgwang@ee, hsli@ee\}.cuhk.edu.hk} \\
\email{\{denghanming, huangyangyi, luotto, daijifeng\}@sensetime.com}\\
}

\maketitle
%\thispagestyle{empty}

%%%%%%%%% ABSTRACT
\begin{abstract}
\begin{spacing}{0.96}
   Video inpainting aims to fill the given spatiotemporal holes with realistic appearance but is still a challenging task even with prosperous deep learning approaches.
   Recent works introduce the promising Transformer architecture into deep video inpainting and achieve better performance.
   However, it still suffers from synthesizing blurry texture as well as huge computational cost.
   Towards this end, we propose a novel Decoupled Spatial-Temporal Transformer (DSTT) for improving video inpainting with exceptional efficiency.
   Our proposed DSTT disentangles the task of learning spatial-temporal attention into 2 sub-tasks: one is for attending temporal object movements on different frames at same spatial locations, which is achieved by temporally-decoupled Transformer block, and the other is for attending similar background textures on same frame of all spatial positions, which is achieved by spatially-decoupled Transformer block. The interweaving stack of such two blocks makes our proposed model attend background textures and moving objects more precisely, and thus the attended plausible and temporally-coherent appearance can be propagated to fill the holes.
   In addition, a hierarchical encoder is adopted before the stack of Transformer blocks, for learning robust and hierarchical features that maintain multi-level local spatial structure, resulting in the more representative token vectors.
   %for better spatial and temporal propagation in follow-up multi-head self-attention modules.
   Seamless combination of these two novel designs forms a better spatial-temporal attention scheme and our proposed model achieves better performance than state-of-the-art video inpainting approaches with significant boosted efficiency.
   Training code and pretrained models are available at \url{https://github.com/ruiliu-ai/DSTT}.
\end{spacing}
\vspace{-1em}
\end{abstract}

%%%%%%%%% BODY TEXT
\section{Introduction}
\label{introduction}
\begin{figure}[t]
    \centering
    \includegraphics[width=1\linewidth]{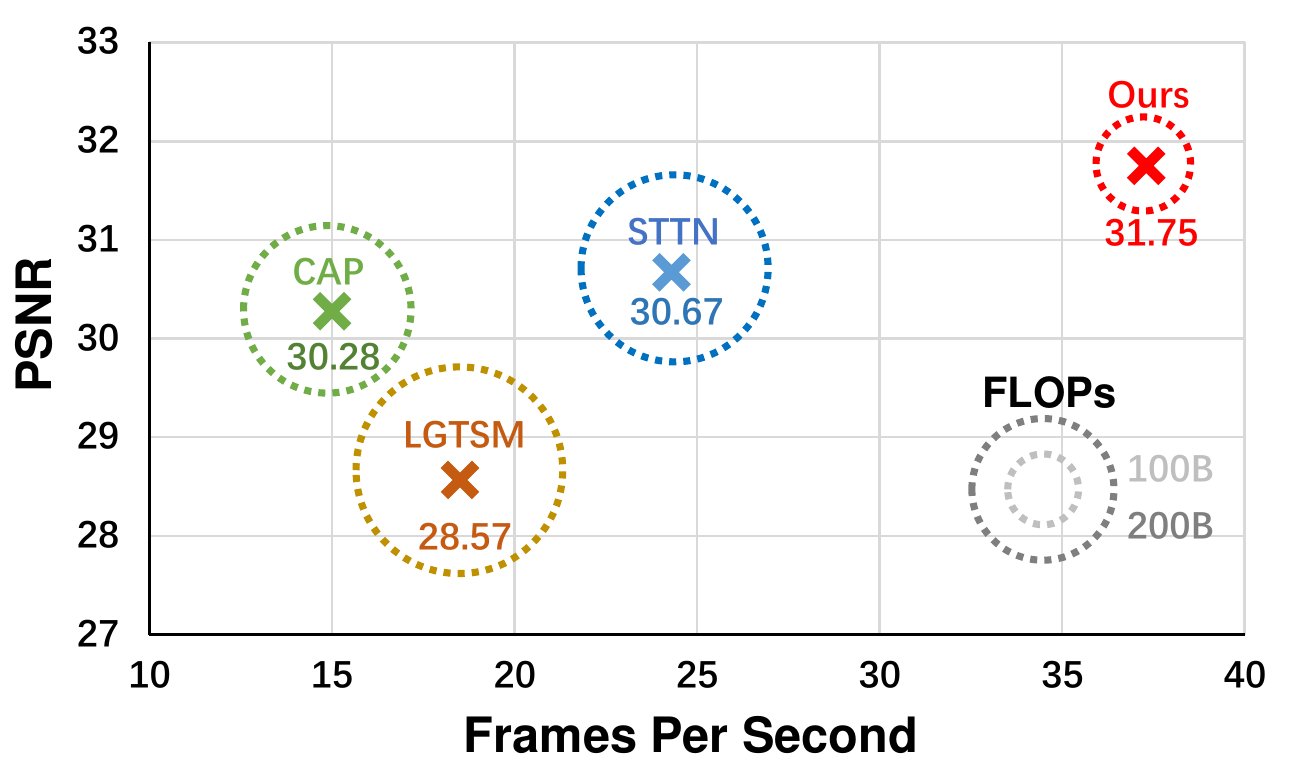}
    \caption{Frames per second vs. PSNR accuracy on different approaches. Our DSTT model achieves the SOTA $31.75$ PSNR value, yet improves the efficiency significantly to FPS $37.3$ with much fewer model FLOPs. FLOPs are indicated by the radius of dashed circle. B is short for Billion. }
    \label{fig:intro}
\vspace{-1em}
\end{figure}

%\color{red}task description.\color{black}
%\begin{spacing}{0.96}
Video inpainting is the task to fill spatiotemporal holes in video frames with realistic and coherent content. It has a wide range of real-world applications such as object removal~\cite{opn, sttn}, video completion~\cite{lgtsm, chang2019free} and video restoration~\cite{cap}.

%\color{red}task challeng/related work/drawbacks.\color{black}
Despite the prevalence of deep learning methods, video inpainting remains a challenging task due to the complexity of visual contents and deteriorated video frames such as motion blur, occlusion induced by camera movement and large object movements. Although deep learning-based image inpainting approaches have made great progress~\cite{iizuka2017globally,  yu2018generative, yu2019free}, processing videos frame-by-frame can hardly generate videos with temporal consistency. Some other works used propagation-based methods to align and propagate temporally-consistent visual contents to obtain more vivid inpainting results~\cite{dfvi, flowedge}. However, they usually fail when faced with occluded stationary backgrounds.
End-to-end training a deep generative model for video inpainting and performing direct inference has drawn great attention in recent years. These methods, however, usually yield blurry videos by using $3D$ convolution~\cite{wang2019aaai, lgtsm, chang2019free}.
Very recently, attention mechanisms are adopted to further promote both realism and temporal consistency via capturing long-range correspondences.
Temporally-consistent appearance is implicitly learned and propagated to the target frame via either frame-level attention~\cite{opn} or patch-level attention~\cite{sttn}.

\begin{figure}[t]
    \centering
    \includegraphics[width=1\linewidth]{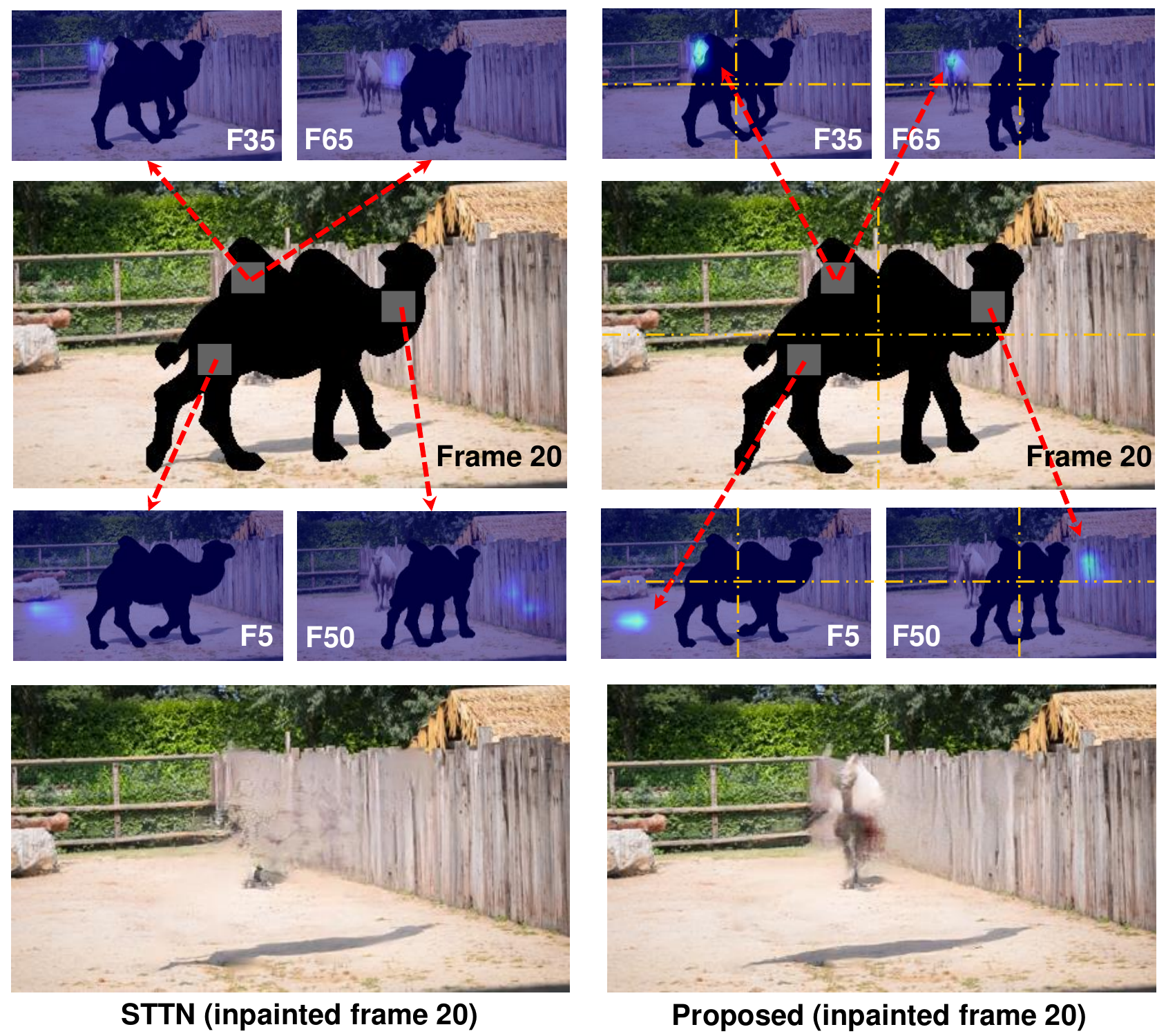}
    \caption{Illustration of the attention maps for missing regions learned by STTN~\cite{sttn} and the proposed DSTT at the first Transformer block. Red lines indicate the attention map of target hole patch on other reference frames. Orange lines indicate the scheme of temporal decoupling that splits frame into 4 zones. With such spatial splitting operation, the self-attention module calculates the attention score among tokens from the same zone of different frames, making this task easier. Therefore, our method can attend occluded object on other reference frames and propagate its appearance to fill the hole more precisely. }
    \label{fig:attn}
\vspace{-1em}
\end{figure}

These methods improve video inpainting performance to a great extent, yet still suffer from the following shortcomings:
On the one hand, processing spatial-temporal information propagation with self-attention modules is challenging in complex scenarios, for example, the arbitrary object movement (see Fig.~\ref{fig:attn}). A camel is occluded for some frames and appears at some other frames, but the previous method~\cite{sttn} can not attend and propagate it precisely.
On the other hand, Transformer-based attention inevitably brings huge computational cost, hindering its usage in real-world applications.

To tackle the challenges, we propose the decoupled spatial-temporal Transformer (DSTT) framework for video inpainting with an interweaving stack of temporally-decoupled and spatially-decoupled Transformer blocks.
Specifically, following STTN~\cite{sttn}, we first extract the downsampled feature maps and embed this feature map into many spatiotemporal token vectors with a convolution-based encoder. Then the obtained spatiotemporal tokens pass through a stack of Transformer blocks for thorough spatial-temporal propagation. Different from the Transformer block in STTN that simultaneously processes spatial-temporal propagation, we decouple the spatial propagation and temporal propagation into $2$ different Transformer blocks.
In temporally-decoupled Transformer block, the obtained tokens are spatially divided into multiple zones, and the model calculates the attention between tokens from different frames' same zone, which is used to model temporal object movement.
%This design is to model temporal object movement, inspired by the basic assumption that temporal movement is usually continuous and thus is easy to be found at neighboring locations of other reference frames.
By contrast, in spatially-decoupled Transformer block, the model calculates the attention between all tokens from the same frame. This is designed for modelling the stationary background texture in the same frame.
These two different Transformer blocks are stacked in an interweaving manner such that all spatial-temporal pixels can be interacted with each other, leading to thorough spatial-temporal searching and propagation. Meanwhile, this spatial-temporal operation becomes much easier after this decoupling scheme, resulting in better video inpainting performance, as shown in Fig.~\ref{fig:attn}. Not only can the occluded object be displayed, but also more coherent texture can be synthesized.

Although an effective and efficient way is provided for thorough spatiotemporal propagation, learning representative and robust feature representation before entering the Transformer blocks is still significant. Because roughly separating image into many patches and squeezing them into token vectors like ViT~\cite{vit} is not a good way to dealing with low-level vision tasks due to the requirement of maintaining spatial structure. Therefore we also design a hierarchical CNN-based encoder for making the extracted feature maps maintain multi-level local spatial structure.
By mixing feature maps from various levels that possess different sizes of receptive fields, the characteristics of obtained feature maps vary from channel to channel, which is beneficial to reaching information interaction along spatial and temporal dimension in following Transformer blocks.
%\end{spacing}

By adopting such a pipeline, the proposed approach achieves state-of-the-art performance on video completion and object removal with much fewer FLOPs and achieves faster inference speed (see Fig.~\ref{fig:intro}).
In summary:
\begin{itemize}
  \item We propose a novel decoupled spatial-temporal Transformer (DSTT) framework for video inpainting to improve video inpainting quality with higher running efficiency, by disentangling the task of spatial-temporal propagation into 2 easier sub-tasks. A hierarchical encoder is also introduced for capturing robust and representative features. The seamless combination of decoupled spatial-temporal Transformer and hierarchical encoder benefits video inpainting performance a lot.

  \item Extensive experiments demonstrate that the proposed DSTT can outperform other state-of-the-art video inpainting approaches both qualitatively and quantitatively, even with significant boosted efficiency.
\end{itemize}

\section{Related Work}

\subsection{Conventional Video Inpainting}
Prior to the prevalence of deep learning, conventional video inpainting approaches usually extend from patch-based image inpainting methods~\cite{10.1145/344779.344972, 10.1109/TIP.2003.815261, ImageMelding12, Efros99texturesynthesis, 10.1145/383259.383296}.
Sampling patches from known region to fill the holes was firstly proposed in~\cite{Efros99texturesynthesis} and developed to reach a milestone as a commercial product~\cite{patchmatch}.
Patwardhan \textit{et al.} extended image texture synthesis~\cite{Efros99texturesynthesis} to temporal domain for video inpainting based on a greedy patch completion scheme ~\cite{occlude, 10.1109/TIP.2006.888343} but these methods only work for stationary camera or constrained camera deformation while ignores the challenge of dynamic camera motion in most natural video sequences.

For solving this problem, Wexler \textit{et al.} directly take $3$-D spatiotemporal patches as processing units and cast this as a global optimization problem by alternating patch search step and color reconstruction step~\cite{10.1109/TPAMI.2007.60}.
Newson \textit{et al.} adapt PatchMatch algorithm~\cite{patchmatch} to video
for further enhancing the temporal dependencies and accelerating the process of patch matching~\cite{Newson2014:SIIMS-VideoInpainting}.
Flow and color completion~\cite{Strobel2014FlowAC} introduce a motion field estimation step for capturing pixel flow more accurately.
Finally Huang \textit{et al.} take all of steps including patch search, color completion and motion field estimation into account simultaneously and perform an alternate optimization on these steps, reaching excellent performance~\cite{huang2016videocompletion}.
Although it works well in various settings, the huge computational complexity during optimization process impedes its wide application in real scenarios.

\subsection{Deep Video Inpainting}
Deep learning boosts image inpainting with excellent performance and high efficiency~\cite{iizuka2017globally, liu2018image, yu2018generative, yu2019free}, which inspires its transition to video modality.
Deep learning-based video inpainting approaches usually fall into propagation-based approaches and inference-based approaches.

On the one hand, propagation-based approaches with optical flow, represented by DFVI~\cite{dfvi} and its edge-aware extension~\cite{flowedge}, use off-the-shelf deep learning-based flow extraction technique~\cite{IMKDB17} for obtaining optical flow with higher robustness and efficiency.
Then they need to learn a deep neural network to complete the obtained flows according to the corrupting mask and finally propagate appearance from reference frames to the target frame for completing video sequences.
However, these methods usually fail when faced with minor background movements in that they can not find relevant texture among those reference frames.
There are other attempts for learning the video completion network with flow or homography estimation module on-the-fly, for performing pixel propagation at the intermediate neural network under the guidance of optical flow~\cite{zhang2019internal}, depth~\cite{DVI} or homography~\cite{cap}.

On the other hand, Wang \textit{et al.} for the first time combine $2$D and $3$D convolution in a fully convolutional neural network for directly inferring missing contents in video sequences~\cite{wang2019aaai}. VINet adopts a recurrent neural network to aggregate temporal features from neighboring frames~\cite{vinet}. Chang \textit{et al.} develop a learnable Gated Temporal Shift Module and improve it to $3$D gated convolution with temporal PatchGAN for free-form video inpainting~\cite{lgtsm, chang2019free}. Hu \textit{et al.} combine $3$D convolution with a region proposal-based strategy for refining inpainting results~\cite{hu2020proposal}.

For better modeling long-range correspondences from distant frames in video sequences, OPN~\cite{opn} proposes an asymmetric attention for calculating the similarities between hole regions on target frame and valid regions on reference frames. Similarly an attention-based dynamic long-term context aggregation module is proposed for globally refining feature maps in~\cite{shortlongterm}. Going a further step, STTN~\cite{sttn} directly transfer the overall multi-head Transformer architecture~\cite{attention} into this task and propose to learn a deep generative model along spatial-temporal dimension. However, it brings a huge computational cost on training and inference, causing the impractical usage in real-time applications. In this work, we propose a novel and effective video inpainting framework endorsed by $2$ carefully-designed modules, which boosts the efficiency significantly.

\begin{figure*}[t]
    \centering
    \includegraphics[width=1\linewidth]{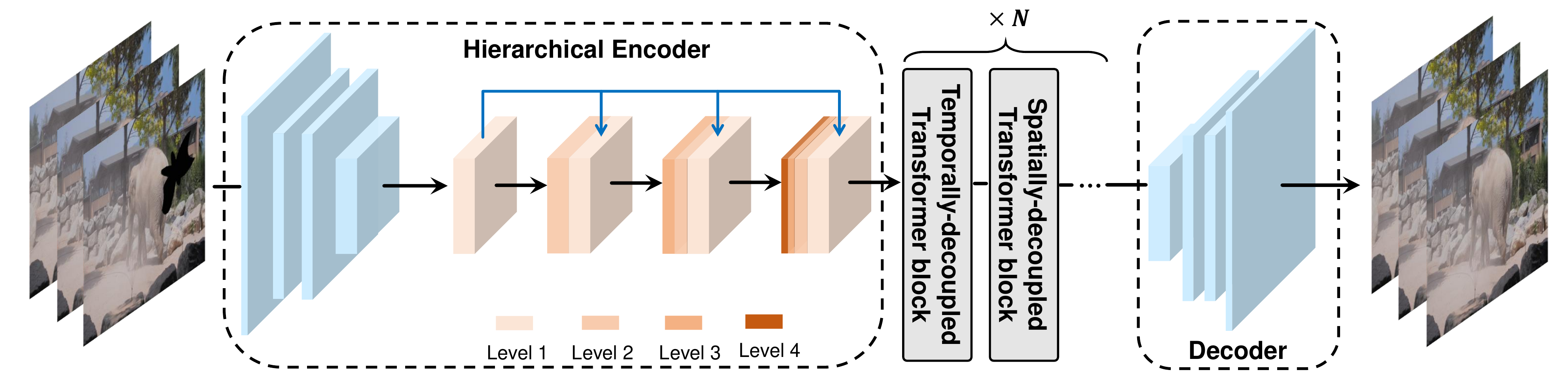}
    \caption{The illustration of proposed framework which is composed of a hierarchical encoder and a stack of decoupled temporal-spatial Transformer blocks, for improving its effectiveness and efficiency. In hierarchical encoder, features of different levels are represented as different colors. The higher the level of feature is, the darker its color is. With such hierarchical grouping processing, the local spatial structures of difference levels are stored in the final robust and representative features. }
    \label{fig:HE}
\vspace{-1em}
\end{figure*}

\section{Method}
Video inpainting targets at filling a corrupted video sequence with plausible contents and smooth pixel flows.
Let $\mathbf{X}^t=\{X^1, X^2, \dots, X^t\}$ denote a corrupted video sequence with length $t$ where $X^i$ means the $i$-th frame of this video sequence.
We aim to learn a mapping function that takes the corrupted video $\mathbf{X}^t$ as input and outputs a realistic video $\mathbf{Y}^t$.
However, there is usually no ground truth $\mathbf{Y}^t$ provided for training in real scenarios.
We therefore formulate this problem as a self-supervised learning framework by randomly corrupting an original videos with a randomly generated mask sequence $\mathbf{M}^t$, \ie, $\mathbf{X}^t = \mathbf{Y}^t \odot \mathbf{M}^t$ where $\odot$ denotes element-wise multiplication.
$\mathbf{M}^i$ is a binary mask whose values are either $0$ referring to the original regions or $1$ meaning the corrupted regions that need filling. Our goal is thus to learn a mapping $G$ for reconstructing the original videos, \textit{i.e.}, $\hat{\mathbf{Y}}^t=G(\mathbf{X}^t, \mathbf{M}^t)$, benefited from which a huge quantity of natural videos can be used for training our framework.

The overall framework of our proposed method is illustrated in Fig.~\ref{fig:HE}.
Given a corrupted video sequence with its corresponding mask sequence, individual frames are first input into a frame-based hierarchical encoder $G_{\text{HE}}$ for downsampling and embedding into spatial-temporal token vectors. With such hierarchical channel-mixing encoder, the robust and representative feature maps are extracted from the input video clip.
After that, we feed the obtained tokens into an interweaving stack of Decoupled Spatial-Temporal Transformer (DSTT) $G_{\text{DSTT}}$ for fully aggregating information across tokens of different frames.
At last, these processed tokens are transformed by a frame-based CNN decoder $G_{\text{Dec}}$ and synthesize the final frames, composing the inpainted video.
The overall generator $G$ can thus be formulated as $$G = G_{\text{HE}}\circ G_{\text{DSTT}} \circ G_{\text{Dec}}.$$

\subsection{Hierarchical Encoder}
\label{HE}
As briefly introduced above, our proposed model includes a stack of specifically-designed Transformer blocks for propagating spatial-temporal information from the known regions outside the hole region at each frame.
However, roughly separating image into many patches and squeezing them into token vectors like ViT~\cite{vit} is not a good way to dealing with low-level vision tasks due to the requirement of maintaining spatial structure.
Although the patch-level self-attention provides a general spatial structure based on the relative position between different patches, the local spatial structure in each patch is corrupted, which is unfavorable for image/video synthesis.
Therefore, a smart and elaborate encoder is needed for extracting robust and representative token vectors.

To achieve this, we propose a novel hierarchical encoder (HE) that mixes the feature maps from different levels frame by frame, to produce tokens containing rich information for each frame.
%Inspired by the design of DEFINE~\cite{define} in Transformer-based language modeling, we propose a novel channel reduction module (CRM). With this module, we reduce the channel of input feature to half in order to save the computational cost of the follow-up Transformer-based feature propagation and aggregation (will be discussed in~\ref{MHST}) without sacrificing inpainting quality.
Specifically, given an input image $X^i \in \mathbb{R}^{h\times w\times 3}$, $i=1,\dots,t$, four cascaded convolutions with kernel size $3,3,3,3$ and stride $2,1,2,1$ make it a down-sampled feature map $F^i_1 \in \mathbb{R}^{h/4\times w/4\times c}$. Starting from this feature, we establish our hierarchical structure for multi-level channel mixing, as shown in Fig.~\ref{fig:HE}.
%the original input feature map would be skip-connected to the input to every recursion layer, establishing a short-cut to its hierarchical entries at multiple levels, as illustrated in Fig.~\ref{fig:CRM}. So the gradient can be back-propagated into the input thoroughly via multiple paths, enjoying a more efficient use of parameters, which can not be achieved by regular convolutions.

We name the feature map $F^i_1$ as the first-level feature map for clearly illustrating our hierarchical structure. It first passes through a convolution layer with kernel size $3\times3$ to obtain the second-level feature map $\hat{F}^i_2 \in \mathbb{R}^{h/4\times w/4\times c}$ whose shape is consistent with first-level feature map while possessing a larger receptive field, \textit{i.e.}, a larger local spatial structure. It is then concatenated with the first-level one, having the feature map $F^i_2 \in \mathbb{R}^{h/4\times w/4\times 2c}$ that include both first-level and second-level local spatial structure.
%, we expand it to a feature map $F_1 \in \mathbb{R}^{h\times w\times 3c}$ with a regular convolution kernel and take these two feature maps as the inputs to the $1$-st layer of CRM.
%DESCRIBE WHAT DOES 1ST LAYER OF CRM DO BEFORE MOVE TO THE 2ND LAYER.

%Then $F^i_2$ will also pass through a convolution layer to obtain $\hat{F}^i_3 \in \mathbb{R}^{h/4\times w/4\times c}$ in which both the first-level feature and second-level feature step into a higher level, becoming second-level and third-level feature with half channel dimension, respectively.
%At this time, the original first-level feature $F^i_1$ is concatenated into it again.
Then at each layer of our hierarchical encoder, all different-level features will go into \textit{next}-level, with growing larger receptive field and a first-level feature is always concatenated to the hierarchical feature map. This process is formulated as
\begin{equation}
\label{eq:he}
\begin{aligned}
  & \hat{F}^i_{j+1} = \text{ReLU}(\text{Conv}(F^i_{j})), \hat{F}^i_{j+1} \in \mathbb{R}^{h/4\times w/4\times c} \\
  & F^i_{j+1} = \text{Concat}(\hat{F}^i_{j+1}, F^i_1), j = 1, ..., L-1,
\end{aligned}
\end{equation}
where Conv denotes a regular convolution operation with kernel size $3\times3$, ReLU denotes the Rectified Linear Unit for modelling non-linear transformation, Concat denotes the concatenation along channel dimension and $L$ denotes the total number of hierarchical layers. We empirically set $L=4$ for our final model and the related ablation study is provided in Section~\ref{ablation}.

With this processing, the finally obtained feature map will include multi-level (from first-level to $L$-th level) local spatial structures of the input image, thus leading to robust and representative feature maps, which is beneficial to the follow-up general patch-level spatial-temporal aggregation.
To further maintain the original local spatial structure, we adopt a group convolution at each layer to separately process feature refinement for various levels. Since the highest-level feature always has least channels, we correspondingly set the number of groups at $j$-th layer to $2^{j-1}$, which avoids feature fusion in the intermediate layers and hence the original local spatial structure at all different levels brings into the last output feature.

After $L$ times recursion, the processed feature would be embedded with a convolution layer with kernel size $7\times7$ and stride $3\times3$ to obtain the final output token feature representation $F^i \in \mathbb{R}^{h/12\times w/12\times 2c}, i=1,\dots,t$.

\subsection{Decoupled Spatial Temporal Transformer}
\label{decouple}
\begin{figure}[t]
    \centering
    \includegraphics[width=1\linewidth]{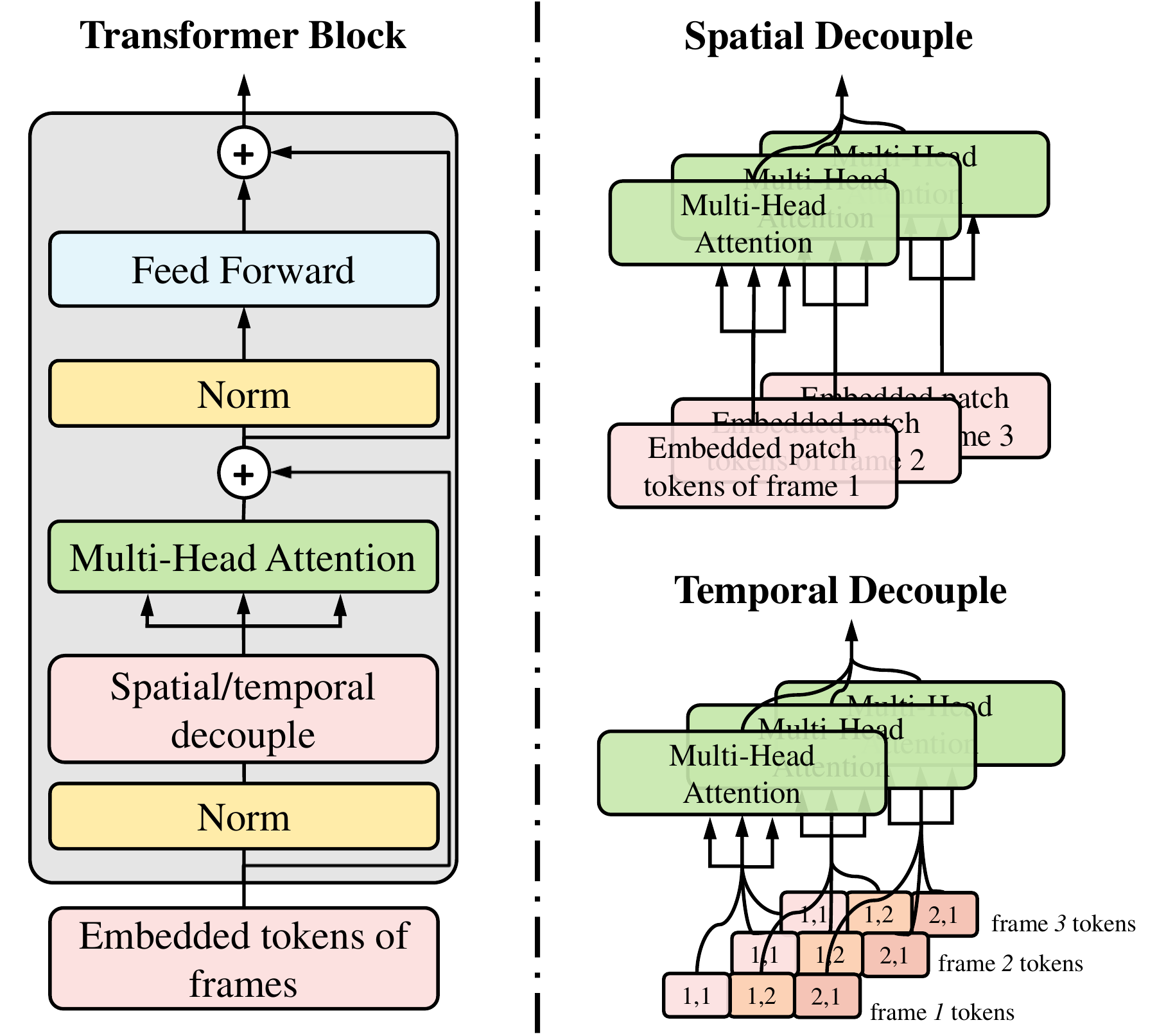}
    \caption{The illustration of a block of the proposed Decoupled Spatial Temporal Transformer (DSTT). A Transformer block consists a decoupled spatial/temporal MSA and a vanilla feed forward net. }
    \label{fig:DSTT}
\vspace{-1em}
\end{figure}
Different from STTN~\cite{sttn} that simultaneously processes spatial and temporal propagation, we disentangle this challenging task into $2$ easier tasks: a temporally-decoupled Transformer block for searching and propagating features from smaller spatial zone along temporal dimension and a spatially-decoupled Transformer block for searching and propagating features from the whole spatial zone without other frames. Two such blocks are stacked for reaching arbitrary pixels across all the spatial-temporal positions.

Specifically, given the token feature $F^i$, we split it into $s^2$ zones along both the height and width dimension where each zone of $F^i$ is denoted as $F_{jk}^i \in \mathbb{R}^{h/(12\cdot s)\times w/(12\cdot s)\times 2c}$ where $j,k = 1,\dots,s$. So far we have a total number of $t\cdot s^2 \cdot n$ tokens where $n=h/(12\cdot s)\cdot w/(12\cdot s)$.
We then take $2$ different ways to grouping these tokens together.
One way is to group them together along temporal dimension, \textit{i.e.}, $P_{jk} = \{F_{jk}^1, \dots, F_{jk}^t\}, j,k=1,\dots,s$ so that temporally-decoupled Transformer block takes each $P_{jk}$ as input and performs multi-head self-attention across tokens in it. By doing so, continuous movement of complex object in a small spatial zone can be detected and propagated into target hole regions.
Another way is to gather them along spatial dimension $P^i = \{F_{11}^i, F_{12}^i, \dots, F_{ss}^i\}, i=1,\dots,t$ so that spatially-decoupled Transformer block takes each $P^{i}$ as input and performs multi-head self-attention across tokens in it. This helps the model detect similar background textures in the spatially-neighboring pixels and propagate these textures to target hole regions to achieve coherent completion.
Two different ways are illustrated clearly as in Fig.~\ref{fig:DSTT}.
Both of temporally-decoupled and spatially-decoupled self-attention modules are followed by a vanilla feed forward network, composing two different Transformer blocks, which is formulated as
\begin{align}
    \label{equ:s}& \hat{P}_{jk} = \text{FFN}(\text{MSA}( P_{jk}) + P_{jk}) + P_{jk}, \\
    \label{equ:t}& \hat{P}^t = \text{FFN}(\text{MSA}( P^t) + P^t) + P^t,
\end{align}
where the former equation stands for the temporally-decoupled Transformer block and the latter stands for the spatially-decoupled Transformer block. FFN denotes the vanilla feed forward net and MSA denotes multi-head self-attention module. Note that the mechanism of temporally-decoupled MSA and spatially-decoupled MSA are same, only their inputs vary in different grouping ways and thus different searching zones.

By doing so, we significantly reduce the computational complexity of MSA from $\mathcal{O}(t^2 s^4 n^2 c)$ into $\mathcal{O}(t^2 s^2 n^2 c)$ (temporally-decoupled MSA) and $\mathcal{O}(t s^4 n^2 c)$ (spatially-decoupled MSA). Although $t=5$ is chosen during training, using a larger $t$ during inference will produce videos with much better visual quality and temporal consistency. So the inference speed is boosted to a great extent.
Following STTN~\cite{sttn}, we set the number of all stacked Transformer blocks to $8$.
As for the stacking strategy, we empirically make a temporally-decoupled block followed by a spatially-decoupled block and repeat this for $4$ times. The related ablation study is conducted in Section~\ref{ablation}.

\begin{figure*}[t]
    \centering
    \includegraphics[width=1\linewidth]{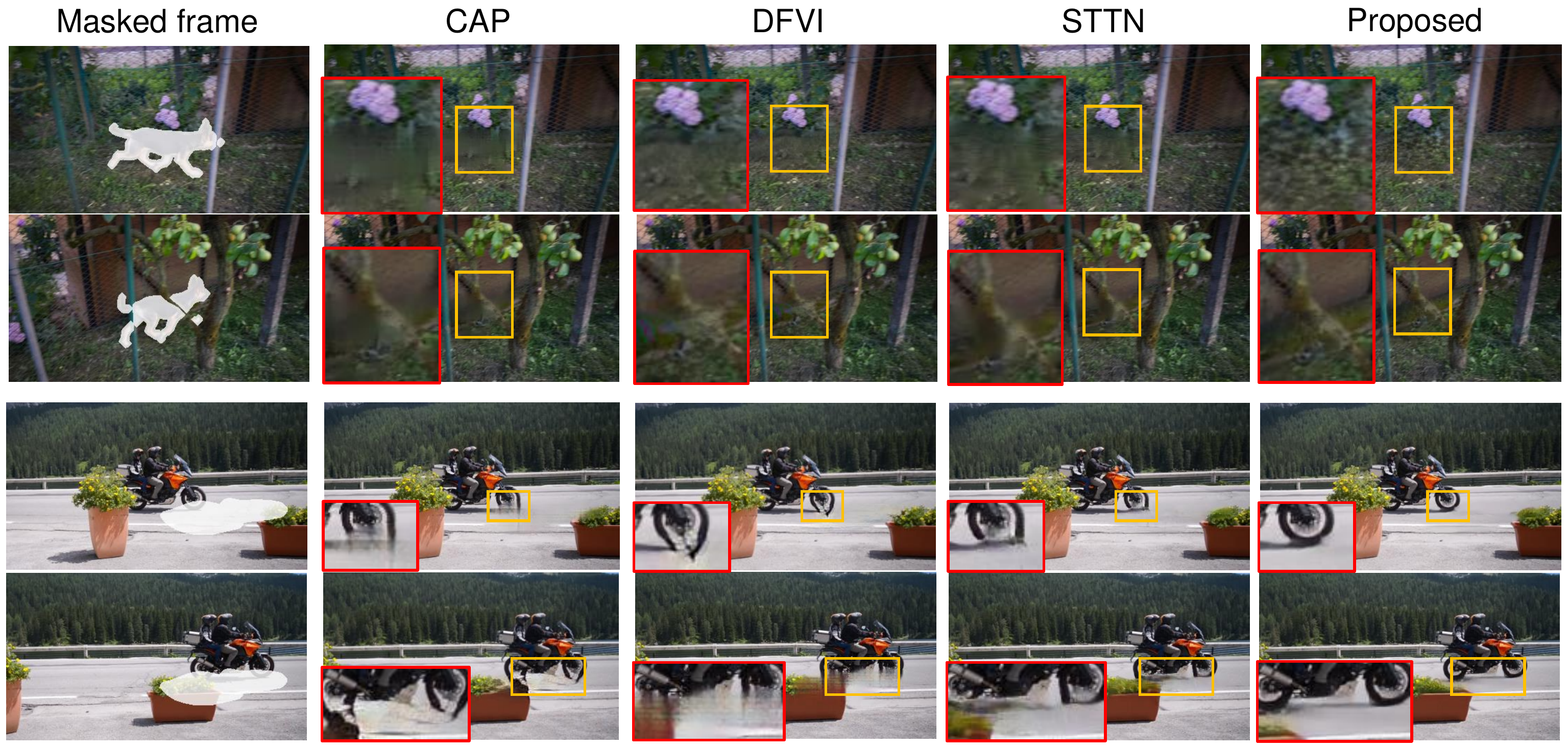}
    \caption{Qualitative comparison with other methods on video completion and object removal. }
    \label{fig:quality}
\vspace{-1em}
\end{figure*}
\subsection{Optimization Objective}
As outlined above, we end-to-end train our DSTT framework in a self-supervised manner.
After passing through the generator $G$ introduced above, we abtain the inpainted video sequence $\hat{\mathbf{Y}}^t=G(\mathbf{X}^t, \mathbf{M}^t)$.

For constraining the synthesized video $\hat{\mathbf{Y}}^t$ to recover the original video $\mathbf{Y}^t$, we firstly choose a simple pixel-level reconstruction loss $\mathcal{L}_1$ between them as our supervision.
Note that the $\mathcal{L}_1$ for valid region and hole region may be of distinct importance so we calculate them separately.
The reconstruction loss for hole region is formulated as:
\begin{equation}
\label{eq:hole}
\begin{aligned}
  \mathcal{L}_\text{hole} = \frac{\|\mathbf{M}^t \odot (\hat{\mathbf{Y}}^t - \mathbf{Y}^t)\|_1}{\|\mathbf{M}^t\|_1}
\end{aligned}
\end{equation}
and the reconstruction loss for valid region is formulated as:
\begin{equation}
\label{eq:valid}
\begin{aligned}
  \mathcal{L}_\text{valid} = \frac{\|(1-\mathbf{M}^t) \odot (\hat{\mathbf{Y}}^t - \mathbf{Y}^t)\|_1}{\|1-\mathbf{M}^t\|_1}
\end{aligned}
\end{equation}
where $\odot$ denotes element-wise multiplication.

In addition, to enhance the realism and temporal coherence of the generated video, we adopt a similar structure of Temporal Patch Generative Adversarial Network (T-PatchGAN)~\cite{chang2019free} as our discriminator $D$, for training with the generator in an adversarial manner.
%The discriminator takes a video as input and extract its spatiotemporal feature representations by cascaded $3$D convolution operations.
The discriminator attempts to distinguish the generated videos from real ones,
%by judging their spatiotemporal features to be $1$ indicating ``real'' or $0$ indicating ``fake''.
while the generator attempts to synthesize a video that would be categorized into ``real'' by the discriminator $D$.
%$D$ and $G$ play a minimax game and finally reach the Nash equilibrium where the discriminator cannot tell the generated video from real one any more.
So the adversarial loss for $D$ is as:
\begin{equation}
\label{eq:advD}
\begin{aligned}
  \mathcal{L}_{D} =  \mathbb{E}_{\mathbf{Y}^t} \left[ \log{D(\mathbf{Y}^t)} \right]
  +  \mathbb{E}_{\hat{\mathbf{Y}}^t} \left[ \log{ (1 - D(\hat{\mathbf{Y}}^t)) } \right]
\end{aligned}
\end{equation}

Contrarily the adversarial loss for $G$ is the opposite direction:
\begin{equation}
\label{eq:advG}
\begin{aligned}
  \mathcal{L}_{\text{adv}} = \mathbb{E}_{\hat{\mathbf{Y}}^t} \left[ \log{ D(\hat{\mathbf{Y}}^t) } \right]
\end{aligned}
\end{equation}

To sum up, our final optimization objective for $G$ is:
\begin{equation}
\label{eq:final}
\begin{aligned}
  \mathcal{L} = \lambda_{\text{hole}} \cdot \mathcal{L}_{\text{hole}} + \lambda_{\text{valid}} \cdot \mathcal{L}_{\text{valid}} + \lambda_{\text{adv}} \cdot \mathcal{L}_{\text{adv}}
\end{aligned}
\end{equation}
where $\lambda_{\text{hole}}$, $\lambda_{\text{valid}}$ and $\lambda_{\text{adv}}$ are hyper-parameters weighing importance for different terms. We empirically set $\lambda_{\text{hole}}$ to $1$, $\lambda_{\text{valid}}$ to $1$ and $\lambda_{\text{adv}}$ to $0.01$ for following experiments.

\section{Experiment}
\subsection{Implementation}
\label{comparison}
\begin{table}
\begin{center}
\caption{\label{table:user}User study on different methods. }
\begin{spacing}{1.1}
\begin{tabular}{l|c|c|c}
\hline
\cline{1-4}
& Rank 1 & Rank 2 & Rank 3 \\
\hline
Ours & $\textbf{47.30}\%$ & $43.22\%$ & $14.44\%$  \\
\hline
CAP~\cite{cap}  & $25.25\%$ & $14.68\%$ & $63.97\%$  \\
\hline
STTN~\cite{sttn}  & $27.45\%$ & $42.10\%$ & $21.59\%$  \\
\hline
\cline{1-4}
\end{tabular}
\end{spacing}
\vspace{-2em}
\end{center}
\end{table}

\begin{table*}\small
\begin{center}
\caption{\label{table:completion}Quantitative results of video completion on YouTube-VOS and DAVIS dataset. B is short for Billion. }
\begin{spacing}{1.19}
\begin{tabular}{l|c|c|c|c||c|c|c|c||c|c}
\hline
\cline{1-11}
& \multicolumn{8}{c||}{Accuracy} & \multicolumn{2}{c}{Efficiency} \\
\cline{2-11}
%\multirow{2}{*}{\backslashbox{Models}{ Metric}}
& \multicolumn{4}{c||}{YouTube-VOS} & \multicolumn{4}{c||}{DAVIS} & \multirow{2}{*}{FLOPs}  & \multirow{2}{*}{FPS}  \\
\cline{1-9}
Models & PSNR $\uparrow$ & SSIM $\uparrow$ & VFID $\downarrow$ &  $\text{E}_{warp} \downarrow$ & PSNR $\uparrow$ & SSIM $\uparrow$ & VFID $\downarrow$ & $ \text{E}_{warp} \downarrow$  & &\\
\cline{1-11}
VINet~\cite{vinet} & 29.20 & 0.9434 & 0.072 & 0.1490 & 28.96 & 0.9411 & 0.199 & 0.1785 & - & - \\
\hline
DFVI~\cite{dfvi}  & 29.16 & 0.9429 & 0.066 & 0.1509 & 28.81 & 0.9404 & 0.187 & 0.1880 & - & - \\
\hline
LGTSM~\cite{lgtsm}  & 29.74 & 0.9504 & 0.070 & 0.1859 & 28.57 & 0.9409 & 0.170 & 0.2566 & 261B & 18.7  \\
\hline
CAP~\cite{cap}  & 31.58 & 0.9607 & 0.071 & 0.1470 & 30.28 & 0.9521 & 0.182 & 0.1824 & 211B &  15.0 \\
\hline
STTN~\cite{sttn}  & 32.34 & \textbf{0.9655} & 0.053 & 0.1451 & 30.67 & 0.9560 & 0.149 & 0.1779 & 233B & 24.3  \\
\hline
Proposed & \textbf{32.66}  & 0.9646  & \textbf{0.052}  & \textbf{0.1430} & \textbf{31.75} & \textbf{0.9650} & \textbf{0.148} & \textbf{0.1716}  & \textbf{128B} & 37.3  \\
\hline
\cline{1-11}
\end{tabular}
\end{spacing}
\vspace{-1em}
\end{center}
\footnotesize{\affmark[1] VINet and DFVI are not end-to-end training methods with complicated stages whose running speed highly rely on optical flow extraction, warpping, etc., so their efficiency are not projectable with others, usually with longer time.}
\vspace{-1em}
\end{table*}

\noindent \textbf{Training details}.
We use Adam optimizer~\cite{adam} with $\beta_1 = 0.9$, $\beta_2 = 0.999$ and its learning rate starts from $0.0001$ and is decayed once with the factor $0.1$ at $400,000$ iteration.
The total training iteration is $500,000$. We use 8 GTX 1080Ti GPUs for training and set batch size to $8$ as well. The total training time is around $50$ hours.

\noindent \textbf{Dataset}.
Two popular video object segmentation datasets are taken for evaluation.
\textit{YouTube-VOS}~\cite{youtubevos} is composed of $3,471$, $474$ and $508$ video clips in training, validation and test set, respectively.
The average number of frames for a video clip is around $150$.
\textit{DAVIS}~\cite{Perazzi2016}, short for Densely Annotated VIdeo Segmentation, contains $150$ video clips with densely annotated object mask sequence.
We split this dataset into a training set including $90$ video clips and a test set including $60$ video clips randomly.
%We set the weight for both valid region and hole region reconstruction loss to 1, and set the weight for adversarial loss to 0.01. We use Adam optimizer with $\beta_1 = 0.9, \beta_2 = 0.999$ and its learning rate starts from $0.0001$ and is decayed once with the factor $0.1$ at $400,000$ iterations. The total training iteration is set to $500,000$.
%For every training video clip, we randomly select $5$ frames from it and resize them to the shape $(432, 240)$, and feed these frames into DSTT framework with corresponding generated masks. Only horizontal flipping is adopted for data augmentation.

\noindent \textbf{Compared methods}.
We mainly choose deep video inpainting approaches for comparison.
\textit{VINet}~\cite{vinet} uses a recurrent neural network to aggregate temporal features for the hole video clip.
\textit{DFVI}\cite{dfvi} propagate relevant appearance from reference frames to target one with completed optical flow.
\textit{CAP}~\cite{cap} performs propagation by a learned homography-based alignment module.
\textit{LGTSM}~\cite{lgtsm} adopts a learnable temporal shift module with a temporal discriminator for completing videos.

\noindent \textbf{Evaluation metrics}.
Structural SIMilarity (\textit{SSIM}) and Peak Signal to Noise Ratio (\textit{PSNR}) are chosen as our metrics for evaluating the quality of reconstructed images compared to original ones.
Higher value of these two metrics indicates better reconstruction quality.
An optical flow-based warping error $E_{warp}$ is chosen for measuring the temporal consistency~\cite{Lai-ECCV-2018}.
%Lower value indicates better temporal consistency.
Video-based Fr\'echet Inception Distance (\textit{VFID}) is adopted for measuring the similarity between the statistics of synthesized videos and that of real videos~\cite{wang2018vid2vid, sttn}.
Lower value suggests perceptually closer to natural videos.
In addition, we count the FLOPs and Params, and record the frame per seconds (FPS) for comparing model efficiency.

%\subsection{Video Completion}
\subsection{Comparison with Other Methods}

We perform experiments by comparing our framework with other deep video inpainting approaches on \textit{video completion} and \textit{object removal}.
For video completion task, we use a sequence of stationary multiple square masks to corrupt the original video clips and wish the algorithm to recover the original videos totally.
For object removal task, we use off-the-shelf moving object segmentation mask sequences to corrupt the corresponding videos and wish the algorithm to synthesize perceptually natural videos with sharp and clear textures.

\noindent \textbf{Qualitative comparison}.
As can be seen in Fig.~\ref{fig:quality}, in video completion task, our framework could synthesize sharp and clear appearance with spatial consistency. In object removal task, our framework could create realistic background textures in invisible or occluded regions. Our model achieves competitive qualitative performance compared with state-of-the-arts on both tasks.

\noindent \textbf{Quantitative comparison}.
Since there is no ground-truth for object removal task, the quantitative evaluation is only conducted on video completion task.
The quantitative results are summarized in Table~\ref{table:completion}, from which we observe that our model obtains best quantitative results in all evaluation metrics on DAVIS dataset. And our model ranks 1-st in \textit{PSNR}, \textit{VFID} and \textit{$E_{warp}$} and ranks 2-nd in \textit{SSIM} on YouTube-VOS test set, yet with fewest FLOPs and highest FPS, showing the strong capability and high efficiency of our model in video inpainting.

\noindent \textbf{User study}.
We further perform a user study to evaluate the visual quality of our model, where CAP~\cite{cap} and STTN~\cite{sttn} are chosen as strong baselines. From \textit{DAVIS}~\cite{Perazzi2016} test set, we randomly sample 15 videos for object removal tasks and 15 videos for video completion. 40 volunteers participate in the user study. For each comparison, we present three videos generated by different approaches and ask volunteers to rank based on visual quality, as summarized in Table~\ref{table:user}.

%\noindent \textbf{Attention map visualization}. We also visualize the attention map of our model at the first Transformer block (Temporally-decoupled) by comparing it with STTN, as shown in Fig.~\ref{fig:attn}. In this figure, red lines indicate the attention map of target hole patch on other reference frames. Orange lines indicate the scheme of spatial splitting operation that split frame into 4 zones, which is served for temporally-decoupled Transformer block. With such spatial splitting operation, the self-attention module calculates the attention score among tokens from the same zone of different frames, making this task easier. Therefore, our method can attend occluded object and propagate its appearance to fill the hole more precisely.

\subsection{Ablation Study}
\label{ablation}
We investigate how each component in our framework impacts the performance of video completion by ablation study. The quantitative metrics \textit{PSNR}, \textit{SSIM} and \textit{VFID} at $250,000$ iteration are summarized in Table~\ref{table:he} to~\ref{table:stack}.
We use \textit{std}, short for standard, to denote our released version of the proposed model, and we use TDB to denote temporally-decoupled block and SDB to denote spatially-decoupled block in following subsections if not specified.

\begin{figure*}[t]
    \centering
    \includegraphics[width=0.9\linewidth]{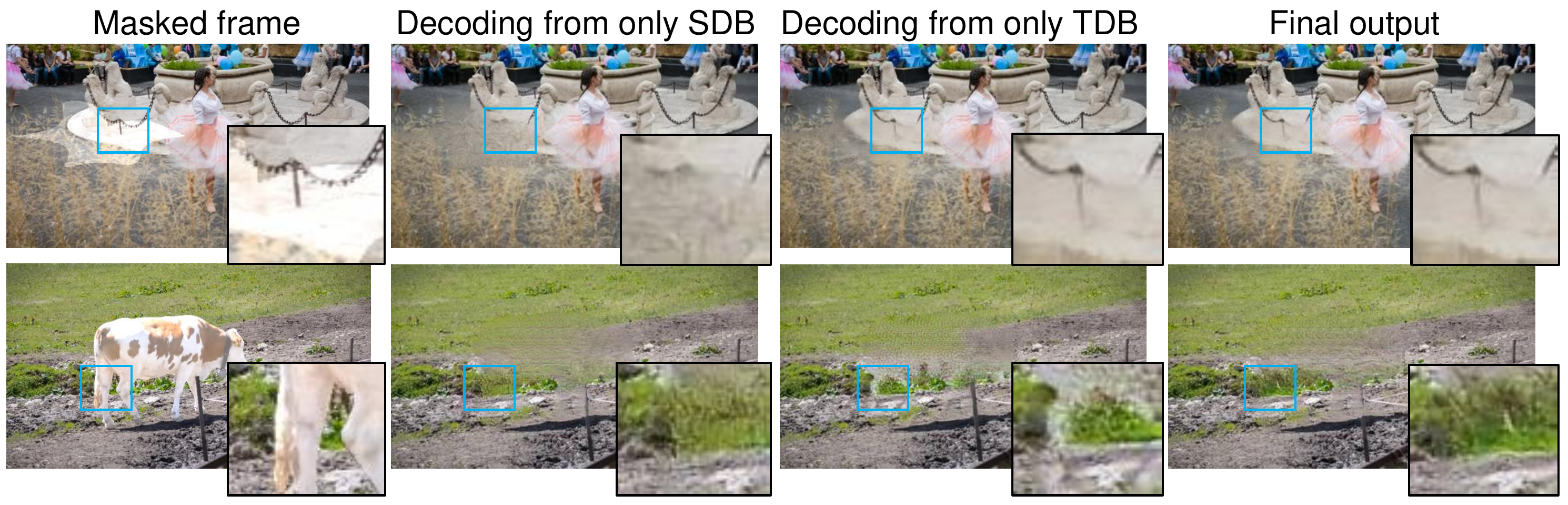}
    \caption{Ablative comparison. The second column shows the result decoded from features only passing through spatially-decoupled blocks (SDB). The third column shows the result decoded from features only passing through temporally-decoupled blocks (TDB). The rightmost column shows the result decoded from features passing through all Transformer blocks. Results from SDB perform better on synthesizing stationary texture and results from TDB perform better on filling the hole with occluded object. The emsembling result hence enjoys better quality and robustness. }
    \label{fig:TSDB}
\vspace{-1em}
\end{figure*}
\noindent \textbf{The effect of TDB and SDB}. We use an ablative comparison by removing the intermediate spatially-decoupled and temporally-decoupled Transformer blocks separately. As shown in Fig.~\ref{fig:TSDB}, the synthesized results from SDB show better stationary texture by detecting and sampling the texture distribution from other pixels in the same frame, and the results from TDB display better appearance of moving object which are propagated from other frames (caused by camera motion in this case). The combination of all SDB and TDB results in better performance and robustness.

\begin{table}
\begin{center}
\caption{\label{table:he}Ablation study on the effect of hierarchical encoder. }
%\vspace{-0.5em}
\begin{spacing}{1.1}
\begin{tabular}{l|c|c|c}
\hline
\cline{1-4}
Models & PSNR $\uparrow$ & SSIM $\uparrow$ & VFID $\downarrow$ \\
\hline
$L=0$  & $30.76$ & $0.9578$ & $0.185$  \\
\hline
$L=1$  & $30.94$ & $0.9593$ & $0.178$  \\
\hline
$L=2$  & $31.05$ & $0.9599$ & $0.175$  \\
\hline
$L=3$  & $31.09$ & $0.9602$ & $0.172$  \\
\hline
$L=4$ (std)  & $31.15$  & $0.9606$ & $0.170$  \\
\hline
$L=5$  & $31.12$ & $0.9604$ & $0.171$  \\
\hline
$L=6$  & $31.16$ & $0.9607$ & $0.170$  \\
\cline{1-4}
\end{tabular}
\end{spacing}
\vspace{-2em}
\end{center}
\end{table}

%\subsubsection{The effect of hierarchical encoder}
%\label{ablation_he}
\noindent \textbf{The effect of hierarchical encoder}.
%\quad We first investigate the effect of hierarchical encoder.
With a hierarchical channel mixing strategy, it learns robust and representative feature maps for follow-up spatial-temporal propagation. We explore its effectiveness of different number of layers on video completion by setting $L=0,\dots,6$.
We first set $L=0$, which means that the original first-level features $F^i_1 \in \mathbb{R}^{h/4\times w/4\times c}$ are directly embedded into multiple tokens for entering the following Transformer blocks.
As can be seen from Table~\ref{table:he}, its \textit{PSNR} and \textit{SSIM} drop by $1.25\%$ and $0.29\%$ respectively, and its \textit{VFID} increases by $8.82\%$, which proves
the importance of learning hierarchical feature representations.
%that without the robust and representative feature learned by our hierarchical encoder, the stack of Transformer blocks fails to propagate useful information along the spatial-temporal dimension.

Then we set $L=1,2,3,5,6$ that lower than the layers of standard model ($L=4$) for investigating the effect of the number of layers adopted in hierarchical encoder on video inpainting performance.
We observe that as the number of layers grows from $1$ to $3$, both \textit{PSNR} and \textit{SSIM} increase step by step, and \textit{VFID} decreases gradually as well, suggesting the effectiveness of our hierarchical design. As the number of layers grows, the aggregated information is more comprehensive from various local spatial structures. However, when it is greater than $5$, the performance does not get better, which indicates that the improvement of such hierarchical scheme is not unlimited under this configuration.
By considering both efficiency and accuracy, we choose $L=4$ as our released version for comparing with state-of-the-arts, as demonstrated in Section~\ref{comparison}.

%\subsubsection{The effect of the number of zones}
\begin{table}
\begin{center}
\caption{\label{table:zone}Ablation study on the effect of the number of zones. }
%\vspace{-0.5em}
\begin{spacing}{1.1}
\begin{tabular}{l|c|c|c|c}
\hline
\cline{1-5}
 & \multicolumn{3}{c|}{Accuracy} & \multirow{2}{*}{FPS} \\
\cline{1-4}
Models    & PSNR $\uparrow$ & SSIM $\uparrow$ & VFID $\downarrow$  \\
\hline
$1 (s=1)$ & $31.10$  & $0.9599$ & $0.172$ & 31.5 \\
\hline
$4 (s=2)$ (std) & $31.15$  & $0.9606$ & $0.170$ & 37.3 \\
\hline
$9 (s=3)$  & $31.05$ & $0.9596$ & $0.170$ & 41.6 \\
\hline
$16 (s=4)$  & $30.92$ & $0.9586$ & $0.171$ & 44.9 \\
\hline
$25 (s=5)$  & $30.82$ & $0.9579$ & $0.171$ & 47.4 \\
\hline
\cline{1-5}
\end{tabular}
\end{spacing}
\vspace{-2em}
\end{center}
\end{table}
%\quad In this part, we investigate the effect of the splitting zones on video inpainting.
\noindent \textbf{The effect of the number of zones}. As mentioned in Section~\ref{decouple}, we need to split the each feature map into $s^2$ zones along the spatial dimension. We now set $s=1,\dots,5$ where the larger $s$ indicates a smaller zone area.
As can be seen from Table~\ref{table:zone}, the model setting $s=1$ that directly searches all spatial-temporal tokens instead of split zones, performs worse than standard model ($s=2$) with \textit{PSNR} and \textit{SSIM} dropped by $0.16\%$ and $0.07\%$ and \textit{VFID} increased by $1.18\%$. This phenomenon verifies our assumption that splitting zones can make the modelling of pixel movements easier. An easier task essentially results in better performance.
In addition, as the value of $s$ grows from $3$ to $5$, the inference speed becomes faster and faster, but the performance drops step by step. Considering both accuracy and efficiency, we finally choose $s=2$ as our released version.

%\subsubsection{The effect of different stacking strategies}
%\label{ablation_stack}

\noindent \textbf{The effect of different stacking strategies}. We investigate the different stacking ways for combining the TDB and SDB, of which the results are reported in Table~\ref{table:stack}.

First, stacking two different blocks in an interweaving manner is always the best option no matter starting from TDB or SDB. By comparing the results in row 1-3 of Table~\ref{table:stack}, we conclude that in the models starting from TDB, the closer the distance between two same blocks is, the model performs worse. Same trend is also found in the models starting from SDB by comparing the row 4-6 of Table~\ref{table:stack}. This is because alternate searching along different dimension directions makes the most flexible way of information propagation.

\begin{table}
\begin{center}
\caption{\label{table:stack}Ablation study on the effect of different stacking strategies. t is short for temporally-decoupled block and s is short for spatially-decoupled block. xn stands for repeating n times. For instance, t1s1-x4 means that temporally-decoupled block is followed by a spatially-decoupled one, which repeats 4 times. }
\begin{spacing}{1.1}
\begin{tabular}{l|c|c|c}
\hline
\cline{1-4}
Models   & PSNR $\uparrow$ & SSIM $\uparrow$ & VFID $\downarrow$   \\
\hline
t1s1-x4 (std)  & $31.15$  & $0.9606$ & $0.170$  \\
\hline
t2s2-x2         & $31.03$  & $0.9596$ & $0.172$ \\
\hline
t4s4-x1         & $30.94$  & $0.9591$ & $0.173$ \\
\hline
s1t1-x4         & $31.06$  & $0.9603$ & $0.173$  \\
\hline
s2t2-x2         & $30.95$  & $0.9592$ & $0.177$ \\
\hline
s4t4-x1         & $30.85$  & $0.9583$ & $0.181$ \\
\hline
\cline{1-4}
\end{tabular}
\end{spacing}
\vspace{-2em}
\end{center}
\end{table}

Then we compare the row 1 and row 4 of Table~\ref{table:stack}, both of which are composed of one TDB followed by one SDB. But the model starting from SDB performs worse than the model starting from TDB. The PSNR and SSIM of s1t1-x4 drops by $0.29\%$ and $0.03\%$, respectively and its VFID is increased by $1.76\%$, compared with our standard t1s1-x4 model. Similar phenomenon could be found in row 2 compared with row 5 and row 3 compared with row 6.
We argue this is because starting from TBD can borrow feature from other frames at the first Transformer block, which utilizes the internal characteristics of videos that pixel flowing is continuous as time grows, making a good starting point.
However, the model starting from SDB can merely searching other patches in its own frame at the first layer, which may not always find enough good texture, especially facing the hard case of occlusion or complex background.

\section{Conclusion}
In this work, we propose a novel DSTT framework for effective and efficient video inpainting. This is achieved by our specifically-designed decoupled spatial-temporal Transformer with hierarchical encoder. Extensive experiments verify that our method achieves state-of-the-art performance in video completion and object removal compared with other video inpainting approaches, with significant boosted efficiency.

{\small
\bibliographystyle{ieee_fullname}

\begin{thebibliography}{10}\itemsep=-1pt

\bibitem{patchmatch}
Connelly Barnes, Eli Shechtman, Adam Finkelstein, and Dan~B Goldman.
\newblock {PatchMatch}: A randomized correspondence algorithm for structural
  image editing.
\newblock {\em ACM Transactions on Graphics (Proc. SIGGRAPH)}, 2009.

\bibitem{10.1145/344779.344972}
Marcelo Bertalmio, Guillermo Sapiro, Vincent Caselles, and Coloma Ballester.
\newblock Image inpainting.
\newblock In {\em Proceedings of the 27th Annual Conference on Computer
  Graphics and Interactive Techniques}, 2000.

\bibitem{10.1109/TIP.2003.815261}
M. Bertalmio, L. Vese, G. Sapiro, and S. Osher.
\newblock Simultaneous structure and texture image inpainting.
\newblock {\em IEEE Transactions on Image Processing}, page 882¨C889, 2003.

\bibitem{chang2019free}
Ya-Liang Chang, Zhe~Yu Liu, Kuan-Ying Lee, and Winston Hsu.
\newblock Free-form video inpainting with 3d gated convolution and temporal
  patchgan.
\newblock {\em In Proceedings of the International Conference on Computer
  Vision (ICCV)}, 2019.

\bibitem{lgtsm}
Ya-Liang Chang, Zhe~Yu Liu, Kuan-Ying Lee, and Winston Hsu.
\newblock Learnable gated temporal shift module for deep video inpainting.
\newblock In {\em BMVC}, 2019.

\bibitem{ImageMelding12}
Soheil Darabi, Eli Shechtman, Connelly Barnes, Dan~B Goldman, and Pradeep Sen.
\newblock {I}mage {M}elding: Combining inconsistent images using patch-based
  synthesis.
\newblock {\em ACM Transactions on Graphics (TOG) (Proceedings of SIGGRAPH
  2012)}, 2012.

\bibitem{vit}
Alexey Dosovitskiy, Lucas Beyer, Alexander Kolesnikov, Dirk Weissenborn,
  Xiaohua Zhai, Thomas Unterthiner, Mostafa Dehghani, Matthias Minderer, Georg
  Heigold, Sylvain Gelly, Jakob Uszkoreit, and Neil Houlsby.
\newblock An image is worth 16x16 words: Transformers for image recognition at
  scale.
\newblock In {\em International Conference on Learning Representations}, 2021.

\bibitem{Efros99texturesynthesis}
Alexei Efros and Thomas Leung.
\newblock Texture synthesis by non-parametric sampling.
\newblock In {\em In International Conference on Computer Vision}, 1999.

\bibitem{10.1145/383259.383296}
Alexei~A. Efros and William~T. Freeman.
\newblock Image quilting for texture synthesis and transfer.
\newblock In {\em Proceedings of SIGGRAPH}.

\bibitem{flowedge}
Chen Gao, Ayush Saraf, Jia-Bin Huang, and Johannes Kopf.
\newblock Flow-edge guided video completion.
\newblock In {\em Proceedings of the European Conference on Computer Vision
  (ECCV)}, 2020.

\bibitem{hu2020proposal}
Yuan-Ting Hu, Heng Wang, Nicolas Ballas, Kristen Grauman, and Alexander~G.
  Schwing.
\newblock Proposal-based video completion.
\newblock In {\em The Proceedings of the European Conference on Computer Vision
  (ECCV)}, 2020.

\bibitem{huang2016videocompletion}
Jia-Bin Huang, Sing~Bing Kang, Narendra Ahuja, and Johannes Kopf.
\newblock Temporally coherent completion of dynamic video.
\newblock {\em ACM Trans. Graph.}, 2016.

\bibitem{iizuka2017globally}
Satoshi Iizuka, Edgar Simo-Serra, and Hiroshi Ishikawa.
\newblock Globally and locally consistent image completion.
\newblock {\em ACM Transactions on Graphics (ToG)}, 2017.

\bibitem{IMKDB17}
E. Ilg, N. Mayer, T. Saikia, M. Keuper, A. Dosovitskiy, and T. Brox.
\newblock Flownet 2.0: Evolution of optical flow estimation with deep networks.
\newblock In {\em IEEE Conference on Computer Vision and Pattern Recognition
  (CVPR)}, Jul 2017.

\bibitem{vinet}
Dahun Kim, Sanghyun Woo, Joon-Young Lee, and In~So Kweon.
\newblock Deep video inpainting.
\newblock In {\em Proceedings of the IEEE conference on computer vision and
  pattern recognition}, 2019.

\bibitem{adam}
Diederik~P. Kingma and Jimmy Ba.
\newblock Adam: {A} method for stochastic optimization.
\newblock In {\em 3rd International Conference on Learning Representations,
  {ICLR}}, 2015.

\bibitem{Lai-ECCV-2018}
Wei-Sheng Lai, Jia-Bin Huang, Oliver Wang, Eli Shechtman, Ersin Yumer, and
  Ming-Hsuan Yang.
\newblock Learning blind video temporal consistency.
\newblock In {\em European Conference on Computer Vision}, 2018.

\bibitem{cap}
Sungho Lee, Seoung~Wug Oh, DaeYeun Won, and Seon~Joo Kim.
\newblock Copy-and-paste networks for deep video inpainting.
\newblock In {\em Proceedings of the IEEE International Conference on Computer
  Vision}, 2019.

\bibitem{shortlongterm}
Ang Li, Shanshan Zhao, Xingjun Ma, Mingming Gong, Jianzhong Qi, Rui Zhang,
  Dacheng Tao, and Ramamohanarao Kotagiri.
\newblock Short-term and long-term context aggregation network for video
  inpainting.
\newblock In {\em ECCV}, 2020.

\bibitem{DVI}
Miao Liao, Feixiang Lu, Dingfu Zhou, Sibo Zhang, Wei Li, and Ruigang Yang.
\newblock Dvi: Depth guided video inpainting for autonomous driving.
\newblock In {\em The Proceedings of the European Conference on Computer Vision
  (ECCV)}, 2020.

\bibitem{liu2018image}
Guilin Liu, Fitsum~A Reda, Kevin~J Shih, Ting-Chun Wang, Andrew Tao, and Bryan
  Catanzaro.
\newblock Image inpainting for irregular holes using partial convolutions.
\newblock In {\em Proceedings of the European Conference on Computer Vision
  (ECCV)}, 2018.

\bibitem{Newson2014:SIIMS-VideoInpainting}
Alasdair Newson, Andr{\'{e}}s Almansa, Matthieu Fradet, Yann Gousseau, and
  Patrick P{\'{e}}rez.
\newblock {Video Inpainting of Complex Scenes}.
\newblock {\em SIAM Journal on Imaging Sciences}, 2014.

\bibitem{opn}
Seoung~Wug Oh, Sungho Lee, Joon-Young Lee, and Seon~Joo Kim.
\newblock Onion-peel networks for deep video completion.
\newblock In {\em Proceedings of the IEEE International Conference on Computer
  Vision}, 2019.

\bibitem{occlude}
Kedar Patwardhan, G. Sapiro, and M. Bertalmio.
\newblock Video inpainting of occluding and occluded objects.
\newblock 2005.

\bibitem{10.1109/TIP.2006.888343}
K.~A. Patwardhan, G. Sapiro, and M. Bertalmio.
\newblock Video inpainting under constrained camera motion.
\newblock {\em Trans. Img. Proc.}, 2007.

\bibitem{Perazzi2016}
F. Perazzi, J. Pont-Tuset, B. McWilliams, L. {Van Gool}, M. Gross, and A.
  Sorkine-Hornung.
\newblock A benchmark dataset and evaluation methodology for video object
  segmentation.
\newblock In {\em Computer Vision and Pattern Recognition}, 2016.

\bibitem{Strobel2014FlowAC}
M. Strobel, Julia Diebold, and D. Cremers.
\newblock Flow and color inpainting for video completion.
\newblock In {\em GCPR}, 2014.

\bibitem{attention}
Ashish Vaswani, Noam Shazeer, Niki Parmar, Jakob Uszkoreit, Llion Jones,
  Aidan~N Gomez, \L~ukasz Kaiser, and Illia Polosukhin.
\newblock Attention is all you need.
\newblock In {\em Advances in Neural Information Processing Systems}, pages
  5998--6008, 2017.

\bibitem{wang2019aaai}
Chuan Wang, Haibin Huang, Xiaoguang Han, and Jue Wang.
\newblock Video inpainting by jointly learning temporal structure and spatial
  details.
\newblock In {\em AAAI}, 2019.

\bibitem{wang2018vid2vid}
Ting-Chun Wang, Ming-Yu Liu, Jun-Yan Zhu, Guilin Liu, Andrew Tao, Jan Kautz,
  and Bryan Catanzaro.
\newblock Video-to-video synthesis.
\newblock In {\em Advances in Neural Information Processing Systems (NeurIPS)},
  2018.

\bibitem{10.1109/TPAMI.2007.60}
Yonatan Wexler, Eli Shechtman, and Michal Irani.
\newblock Space-time completion of video.
\newblock {\em IEEE Trans. Pattern Anal. Mach. Intell.}, 2007.

\bibitem{youtubevos}
Ning Xu, Linjie Yang, Yuchen Fan, Dingcheng Yue, Yuchen Liang, Jianchao Yang,
  and Thomas Huang.
\newblock Youtube-vos: A large-scale video object segmentation benchmark.
\newblock {\em arXiv: 1809.03327}, 2018.

\bibitem{dfvi}
Rui Xu, Xiaoxiao Li, Bolei Zhou, and Chen~Change Loy.
\newblock Deep flow-guided video inpainting.
\newblock In {\em Proceedings of the IEEE conference on computer vision and
  pattern recognition}, 2019.

\bibitem{yu2018generative}
Jiahui Yu, Zhe Lin, Jimei Yang, Xiaohui Shen, Xin Lu, and Thomas~S Huang.
\newblock Generative image inpainting with contextual attention.
\newblock In {\em Proceedings of the IEEE conference on computer vision and
  pattern recognition}, 2018.

\bibitem{yu2019free}
Jiahui Yu, Zhe Lin, Jimei Yang, Xiaohui Shen, Xin Lu, and Thomas~S Huang.
\newblock Free-form image inpainting with gated convolution.
\newblock In {\em Proceedings of the IEEE International Conference on Computer
  Vision}, 2019.

\bibitem{sttn}
Yanhong Zeng, Jianlong Fu, and Hongyang Chao.
\newblock Learning joint spatial-temporal transformations for video inpainting.
\newblock In {\em The Proceedings of the European Conference on Computer Vision
  (ECCV)}, 2020.

\bibitem{zhang2019internal}
Haotian Zhang, Long Mai, Ning Xu, Zhaowen Wang, John Collomosse, and Hailin
  Jin.
\newblock An internal learning approach to video inpainting.
\newblock In {\em Proceedings of the IEEE International Conference on Computer
  Vision}, 2019.

\end{thebibliography}

}

\end{document}